\documentclass[a4paper,conference]{IEEEtran}
\IEEEoverridecommandlockouts

\usepackage{tikz}
\usetikzlibrary{positioning,arrows.meta,fit,backgrounds,calc,shapes.geometric}
\usepackage{cite}
\usepackage{amsmath,amssymb,amsfonts}
\usepackage{algorithmic}
\usepackage{graphicx}
\usepackage{textcomp}
\usepackage{xcolor}
\usepackage{placeins}
\usepackage{pgfplots}\pgfplotsset{compat=1.18}
\def\BibTeX{{\rm B\kern-.05em{\sc i\kern-.025em b}\kern-.08em
    T\kern-.1667em\lower.7ex\hbox{E}\kern-.125emX}}

\usepackage{xspace}

\usepackage{url}
\usepackage{hyperref}
\usepackage{float}
\usepackage{subcaption}
\usepackage{amsmath}

\DeclareMathAlphabet\mathbfcal{OMS}{cmsy}{b}{n}
\usepackage{bm}

\usepackage{color,array}
\usepackage{xcolor,colortbl}

\definecolor{Gray}{gray}{0.85}
\newcolumntype{a}{>{\columncolor{Gray}}c}

\usepackage[acronym]{glossaries}

\newacronym{dnn}{DNN}{Deep Neural Network}

\newacronym{uap}{UAP}{Universal Adversarial Perturbation}

\newacronym{llm}{LLM}{Large Language Models}

\newacronym{fgsm}{FGSM}{Fast Gradient Sign Method}

\newacronym{pgd}{PGD}{Projected Gradient Descent}

\newacronym{rmse}{RMSE}{Root Mean Squared Error}

\newacronym{mse}{MSE}{Mean Square Error}

\newacronym{ros}{ROS}{Robot Operating System}

\newacronym{nps}{NPS}{Non-Printability Score}

\newacronym{snps}{SNPS}{Sub-sampled Non-Printability Score}

\newacronym{zoo}{ZOO}{Zeroth Order Optimization}

\newacronym{simba}{SimBA}{Simple Black-box Attack}

\newacronym{mad}{MAD}{Mean Absolute Deviation}

\newacronym{pdp}{PDP}{Partial Dependence Plot}

\newacronym{ice}{ICE}{Individual Conditional Expectation}

\newacronym{ale}{ALE}{Accumulated Local Effects}

\newacronym{pfi}{PFI}{Permutation Feature Importance}

\newacronym{lime}{LIME}{Local Interpretable Model-agnostic Explanations}

\newacronym{shap}{SHAP}{Shapley Additive Explanation}

\newacronym{tbd}{TBD}{Tracking-By-Detection}

\newacronym{sot}{SOT}{Single-Object Tracking}

\newacronym{mot}{MOT}{Multi-Object Tracking}

\newacronym{jdt}{JDT}{Joint Detection and Tracking}

\newacronym{hota}{HOTA}{Higher Order Tracking Accuracy}

\newacronym{mt}{MT}{Mostly Tracked}

\newacronym{ml}{ML}{Mostly Lost}

\newacronym{mota}{MOTA}{Multiple Object Tracking Accuracy}

\newacronym{idf1}{IDF1}{Identification F1 Score}

\usepackage{pgfplots}
\pgfplotsset{compat=1.18}

\begin{document}

\title{Adversarial Tracking: Attacks on Image Sensors for Object Tracking in Autonomous Vehicles}


\author{\IEEEauthorblockN{Han Wu, Alex Chan, Ekin Can Erkus, Tousif Rahman and Rishad Shafik}
\IEEEauthorblockA{School of Engineering, Newcastle University, Newcastle upon Tyne, United Kingdom \\
\textit{Corresponding author: han.wu2@newcastle.ac.uk}}}    

\maketitle

\begin{abstract}
Object Tracking systems estimate the motion of surrounding vehicles by linking detected objects across video frames. In particular, Tracking-By-Detection (TBD) systems combine object detectors with motion prediction and data association to maintain object identities over time. However, neural network-based detectors are vulnerable to adversarial attacks that perturb input images, and their impact on downstream trajectory estimation is not fully characterised by detection accuracy alone. Therefore, this paper evaluates image-specific and image-agnostic attacks on YOLOv3 and YOLOv4 detectors paired with SORT, Deep SORT, Strong SORT, and OC-SORT tracking systems. The evaluation uses a CARLA autonomous driving simulator and measures tracking quality through Multiple Object Tracking Accuracy (MOTA), Higher Order Tracking Accuracy (HOTA) and Identification F1 (IDF1). Under image-specific attacks, HOTA for YOLOv3-based trackers declines from 53.45--56.19\% to 14.85--17.03\%, while MOTA reaches -772.60\% in the worst case. YOLOv4-based trackers exhibit greater robustness, while HOTA still declines to 25.06–26.86\%. In addition, image-agnostic attacks can reduce the tracking accuracy using a single Universal Adversarial Perturbation (UAP), without generating a separate perturbation for individual frames.

\end{abstract}



\begin{IEEEkeywords}
Edge Sensors, Autonomous Vehicles, Object Tracking, Adversarial Attacks
\end{IEEEkeywords}

\section{Introduction}
\label{sec:adv_track}
Object Tracking is a fundamental perception task in autonomous driving that estimates the locations and identities of surrounding objects over time~\cite{sun2020survey}. Camera-based object tracking includes \acrfull{sot} and \acrfull{mot}. SOT follows one specified target, whereas MOT follows several objects and maintains a separate tracklet for each~\cite{krebs2017survey}. 

\begin{figure}[tb]
    \centering
    \includegraphics[width=0.48\textwidth]{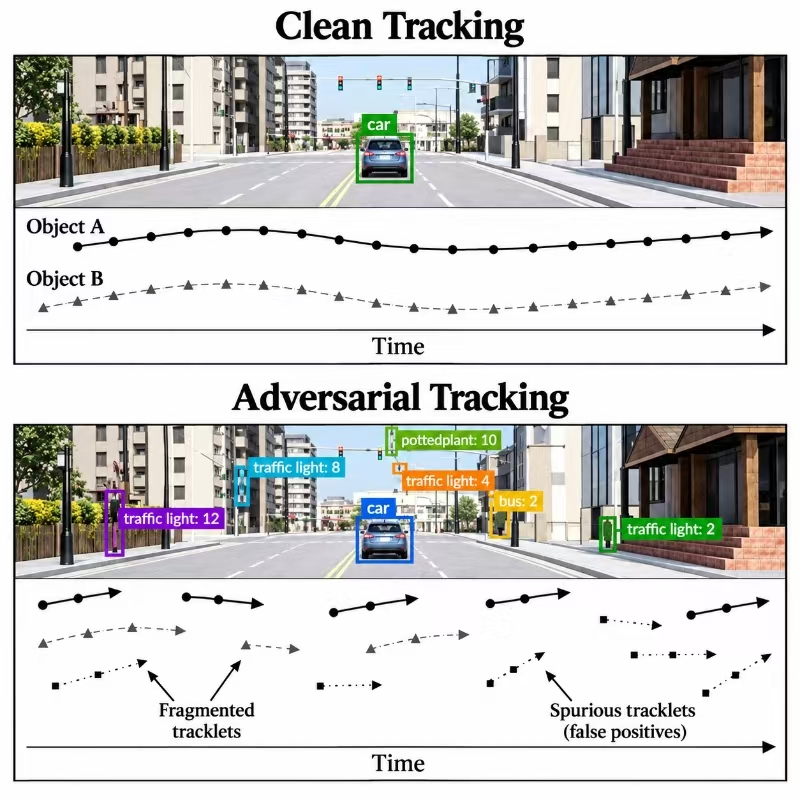}
  \caption{Adversarial Attacks on Object Tracking: Adversarial perturbations disrupt Multi-Object Tracking (MOT) by fragmenting continuous tracklets and generating spurious tracklets.}
  \label{fig:overview}
\end{figure}

Although deep neural networks have substantially advanced the tracking performance, they are vulnerable to adversarial attacks that perturb input images~\cite{jia2020fooling}. For example, \acrfull{tbd} systems combine object detectors with motion prediction and association across consecutive frames. Some \acrshort{tbd} systems rely on pre-trained detection models, and perturbations generated against widely used object detection models, such as YOLO\cite{redmon2018yolov3, bochkovskiy2020yolov4, ge2021yolox}, may transfer to tracking systems. As a result, the tracking performance of \acrshort{tbd} systems heavily relies on the accuracy of object detectors, which is susceptible to adversarial attacks. For example, persistent detection errors can disrupt data association and lead to fragmented trajectories~\cite{lin2021tracklet}, as shown in Fig. \ref{fig:overview}. However, detection errors do not necessarily translate directly into tracking errors, as the data association stage may reject inconsistent detections~\cite{jia2020fooling}.

Therefore, a systematic evaluation at the tracking level is necessary to characterize the impact of adversarial attacks on autonomous driving. The contributions are as follows:


\begin{itemize}
    \item \textbf{CARLA Dataset:} An open-source framework is developed to generate synthetic autonomous driving dataset using CARLA Simulator~\cite{Dosovitskiy17}.\footnote{The CARLA autonomous driving dataset generation code: \url{https://github.com/wuhanstudio/carla-tracking-dataset}.}
    \item \textbf{Tracking Benchmark:}  A benchmark is established to evaluate representative real-time tracking algorithms, including SOT\cite{bewley2016simple}, Deep-SORT\cite{wojke2017simple}, Strong-SORT \cite{du2023strongsort}, and OC-SORT\cite{cao2023observation}, in combination with three object detectors: YOLOv3, YOLOv4, and YOLOX.
    \item \textbf{Adversarial Attacks:} Image-specific and image-agnostic attacks against \acrshort{tbd} systems are evaluated using MOTA, IDF1, and HOTA.\footnote{The source code of Adversarial Attacks is open-sourced on GitHub: \url{https://github.com/wuhanstudio/adversarial-tracking}.}
\end{itemize}


\clearpage

\begin{figure*}[tb]
    \centering
    \includegraphics[width=\textwidth]{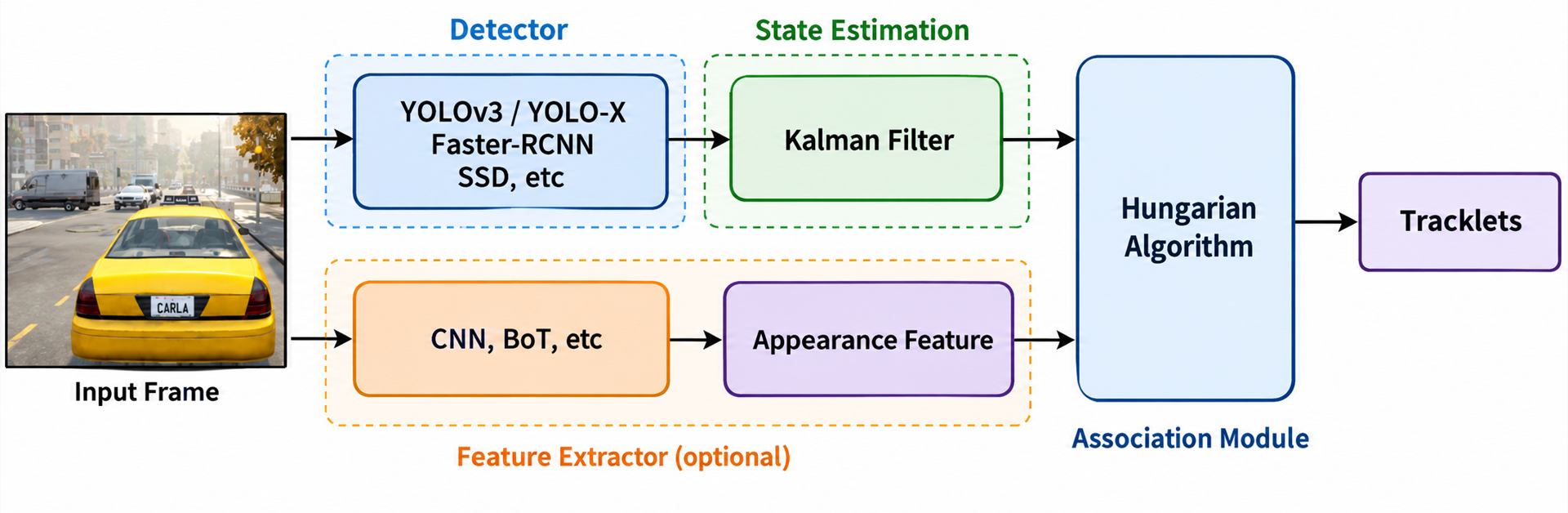}
  \caption{Common components of online tracking methods based on the \acrshort{tbd} framework.}
  \label{fig:tracking_module}
\end{figure*}

\section{Background}

\subsection{Object Tracking}
Camera-based \acrlong{mot} (\acrshort{mot}) methods generally follow two approaches, \acrfull{jdt} and \acrfull{tbd}. \acrshort{jdt} methods integrate detection and tracking, whereas \acrshort{tbd} methods use a modular framework with an object detector, state estimation, an optional feature extractor, and data association~\cite{sun2020survey}. Figure~\ref{fig:tracking_module} summarizes these components and their roles in online tracking.



\noindent\textbf{Object Detector:} The object detector identifies objects in each camera frame and outputs bounding boxes, confidence scores, and predicted classes. Each object will later be assigned a unique and consistent ID across frames.

\noindent\textbf{State Estimation}:  To assign a consistent ID to the same object across frames, SORT-like trackers commonly use a Kalman filter to predict the next state of each track~\cite{bewley2016simple}. Under a linear constant-velocity model, the state vector of each detected object is represented by
\begin{equation}
    \bm{x} = [u, v, s, r, \dot{u}, \dot{v}, \dot{s}]^T
\end{equation}
where $(u, v)$ is the box center, $s$ is its scale, $r$ is its aspect ratio, and $[\dot{u},\dot{v},\dot{s}]$ denotes the corresponding velocities of $u$, $v$ and $s$ respectively. The state vector is updated if a predicted bounding box is associated with the object.



\noindent\textbf{Feature Extractor (Optional)}: Motion state vectors alone may be insufficient to distinguish nearby or occluded objects. Therefore, appearance-based trackers, including Deep SORT, use a Convolutional Neural Network (CNN) with Bag of Tricks (BoT) to extract feature vectors to differentiate objects~\cite{wojke2017simple}. This optional stage is shown in Fig.~\ref{fig:tracking_module}. 

\noindent\textbf{Data Association:} The data association module matches new detections to existing tracks using their predicted states and, optionally, feature vectors extracted from images. The Mahalanobis distance~\cite{chandra1936generalised} can measure the discrepancy between a new detection and a predicted track state. Subsequently, the Hungarian algorithm \cite{kuhn1955hungarian} solves a linear assignment problem to determine the optimal set of detection-to-track matches, maintaining consistent IDs across video frames.


\subsection{Evaluation Metrics}
\label{sec:tracking_eval}
The performance of MOT methods is commonly evaluated using MOTA~\cite{bernardin2008evaluating}, IDF1~\cite{ristani2016performance}, and HOTA~\cite{Luiten2020IJCV}. These metrics capture different aspects of tracking accuracy, including detector errors, identity preservation, and the overall quality of object tracking.



\textbf{Multiple Object Tracking Accuracy (MOTA):} Evaluates the tracking accuracy by considering False Positives (FP), False Negatives (FN), and Identity Switches (IDSW), which reflects both detection and association errors over time $t$. A higher MOTA indicates fewer detection and identity errors:

\begin{equation}
    MOTA =
    1 -
    \frac{\sum_t FN_t + \sum_t FP_t + \sum_t IDSW_t}
    {\sum_t GT_t}
\end{equation}

\textbf{Identification F1 Score (IDF1):} IDF1 measures whether predicted IDs remain consistent with ground-truth identities across the sequence~\cite{ristani2016performance}. It is defined through identification true positives (IDTP), false positives (IDFP), and false negatives (IDFN). A higher IDF1 indicates more consistent identity assignment:

\begin{equation}
    \text{IDF}_1 =
    \frac{2\text{IDTP}}
    {2\text{IDTP}+\text{IDFP}+\text{IDFN}}
\end{equation}

\textbf{Higher Order Tracking Accuracy (HOTA):} Provides a balanced evaluation of Detection Accuracy (DetA) and Association Accuracy (AssA) with different $\alpha$ thresholds ~\cite{Luiten2020IJCV}:

\begin{equation}
\text{HOTA}_{\alpha}=\sqrt{\text{DetA}_{\alpha}\cdot\text{AssA}_{\alpha}}.
\end{equation}

DetA measures how accurately objects are detected, while AssA measures how consistently their identities are maintained across frames. This means that HOTA provides a more balanced assessment of detection and association performance.

Overall, MOTA emphasizes detection and identity errors, IDF1 focuses on identity consistency, and HOTA provides a balanced measure of detection and association performance.

\subsection{Adversarial Attacks}
Both \acrshort{jdt} and \acrshort{tbd} methods are vulnerable to adversarial attacks. For example, end-to-end deep learning models used for \acrshort{jdt}, such as GOTURN~\cite{wiyatno2019physical} and FairMOT~\cite{lin2021trasw}, can be affected by adversarial perturbations, while for \acrshort{tbd} methods, adversaries can target different components of the tracking pipeline, including the Object Detector, Data Association module, and Tracklet Management module.

\textbf{Attacking Object Detectors}: The accuracy of \acrshort{tbd} methods heavily relies on the object detector's ability to produce precise bounding boxes of target objects. Missed or misplaced detections can therefore disrupt subsequent track association. 

In 2019, the first adversarial attack against \acrshort{mot} models attacked a \acrshort{tbd} method that used YOLOv3 as the object detector ~\cite{jia2020fooling}. Moreover, YOLOX ~\cite{ge2021yolox} and CenterNet ~\cite{duan2019centernet} stand out as two widely-used object detectors for autonomous driving applications, yet they both exhibit vulnerability to adversarial attacks ~\cite{zhou2023f,pang2024blinding}. To mitigate adversarial attacks, temporal outlier detection offers a potential approach to identifying adversarial perturbations~\cite{erkus2021fod}.



\textbf{Attacking Association Methods}: Assigning a unique ID to each detected object is crucial for Object Tracking. To attack the association method, Blind Attack assigns the background label to the target object, while the Blur Attack assigns incorrect labels to the target object ~\cite{liu2022efficient}.


\textbf{Attacking Tracklets}: Tracklets, commonly managed by the Kalman Filter, store the position and velocity of each vehicle. By perturbing objects' speed and direction, incorrect states are predicted for each object, enabling the switching of tracklets between adjacent objects. For instance, Lin et al. ~\cite{lin2021tracklet} perturbed the appearance and distance similarity of two adjacent pedestrians, causing FairMOT and ByteTrack models to fail to track the targets in the subsequent frames.

\section{Methodology}

Although some \acrshort{mot} systems use 3D LiDAR measurements\cite{liu2022imperceptible}, this study focuses on vision-only \acrshort{tbd} systems. Specifically, we focus on modular \acrshort{tbd} methods to investigate whether traditional statistical tracking components, such as the Kalman filter, can help mitigate the effects of adversarial attacks. 


Given an input image $x$, the object detection model produces a set of $N$ candidate detections
$\mathcal{H}={h_1,h_2,\ldots,h_N}$. Each detection hypothesis is represented as
\begin{equation}
h_i = (b_i,c_i,\mathbf{p}_i),
\qquad 1\leq i\leq N,
\end{equation}
where $b_i=(b_i^x,b_i^y,b_i^w,b_i^h)$ denotes x, y coordinates, width and height of the predicted bounding box, $c_i\in[0,1]$ is the confidence score, and
$\mathbf{p}_i=(p_i^1,p_i^2,\ldots,p_i^K)$ is the class-probability vector for $K$ object classes.

For an image $x$ at time $t$, the detector therefore produces
\begin{equation}
\mathcal{H}(x_t)
=
{h_1^t,h_2^t,\ldots,h_N^t}.
\end{equation}
Each hypothesis $h_i^t=(b_i^t,c_i^t,\mathbf{p}_i^t)$ describes a candidate object detected in the corresponding frame.


The adversarial attack aims to generate a perturbation $\delta\in[-1,1]^{whc}$ such that the detector produces a different set of hypotheses for the perturbed image
$x'=x+\delta$, i.e.,
\begin{equation}
\mathcal{H}(x')\neq\mathcal{H}(x).
\end{equation}


{\textbf{Image-Specific Attack}}: The image-specific attack generates a perturbation tailored to a single input image. The attack maximizes or minimizes an adversarial loss $\mathcal{L}_{adv}$ with respect to the input image, rather than updating the model parameters. Maximizing the adversarial loss $\mathcal{L}_{adv}$ encourages the detector to generate incorrect bounding boxes, while minimizing them suppresses detected objects. The adversarial perturbation $\delta_t$ at time step $t$ is updated using the following equation:

\begin{equation}
\delta_{t+1}
\leftarrow
\operatorname{Proj}_{\infty,\epsilon}
\left(
\delta_t+
\alpha_t
\operatorname{sign}
\left(
\nabla_x \mathcal{L}_{adv}(x+\delta_t)
\right)
\right)
\end{equation}
where $\alpha_t$ is the learning rate and $\operatorname{Proj}_{\infty,\epsilon}$ constrains the perturbation within an $l_\infty$ bound of the attack strength $\epsilon$. The learning rate is decayed after each iteration.

\textbf{Image-Agnostic Attack}: The same optimization principle can be extended to an image-agnostic attack, which learns a single perturbation that can be applied to multiple input images. Given a set of sample images $X_S$, the perturbation is iteratively optimized by computing the adversarial gradient for each image and accumulating the resulting updates. A smaller learning rate is used to improve the transferability of the perturbation across different images:
\begin{equation}
\delta
\leftarrow
\operatorname{Proj}_{\infty,\epsilon}
\left(
\delta+
\alpha
\operatorname{sign}
\left(
\nabla_x\mathcal{L}_{adv}(x+\delta)
\right)
\right),
\quad x\in X_S.
\end{equation}
Unlike the image-specific attack, which produces a different perturbation for each image, the image-agnostic attack produces a single universal perturbation that is optimized across the available sample images.

Lastly, the adversarial loss $\mathcal{L}_{adv}$ is defined as
\begin{equation}
\mathcal{L}_{adv}(x)
=
\frac{
\displaystyle\sum_{i=1}^{N}
\sigma(c_i)*\sigma(p_i)
}{
\displaystyle\sum_{i=1}^{N}
\left[
\sigma(b_i^w)*\sigma(b_i^h)
\right]^2
},
\end{equation}
where $N$ is the number of candidate boxes, $c_i$ is the confidence score, $p_i$ is the class probability, and $b_i^w$ and $b_i^h$ are the predicted box dimensions. The sigmoid function is denoted by $\sigma(\cdot)$. Thus, the adversarial loss increases the detection confidence for all objects while reducing the size of predicted bounding boxes, thereby increasing the number of detections.


\section{Experimental Results}
\label{sec:results}

The experiment applies image-specific and image-agnostic attacks to SORT~\cite{bewley2016simple}, Deep SORT~\cite{wojke2017simple}, Strong SORT~\cite{du2023strongsort}, and OC-SORT~\cite{cao2023observation} with YOLOv3 and YOLOv4 detectors  in CARLA simulator~\cite{Dosovitskiy17} across seven maps. The evaluation investigates whether conventional tracking components, such as the Kalman filter and the Hungarian algorithm, can mitigate the effects of adversarial attacks.




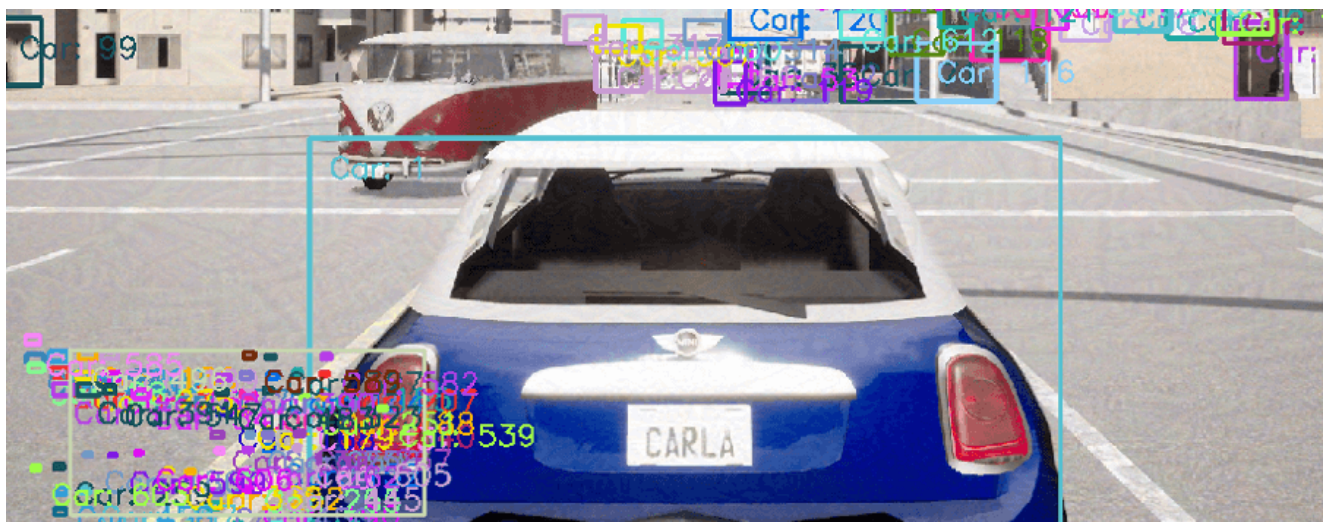
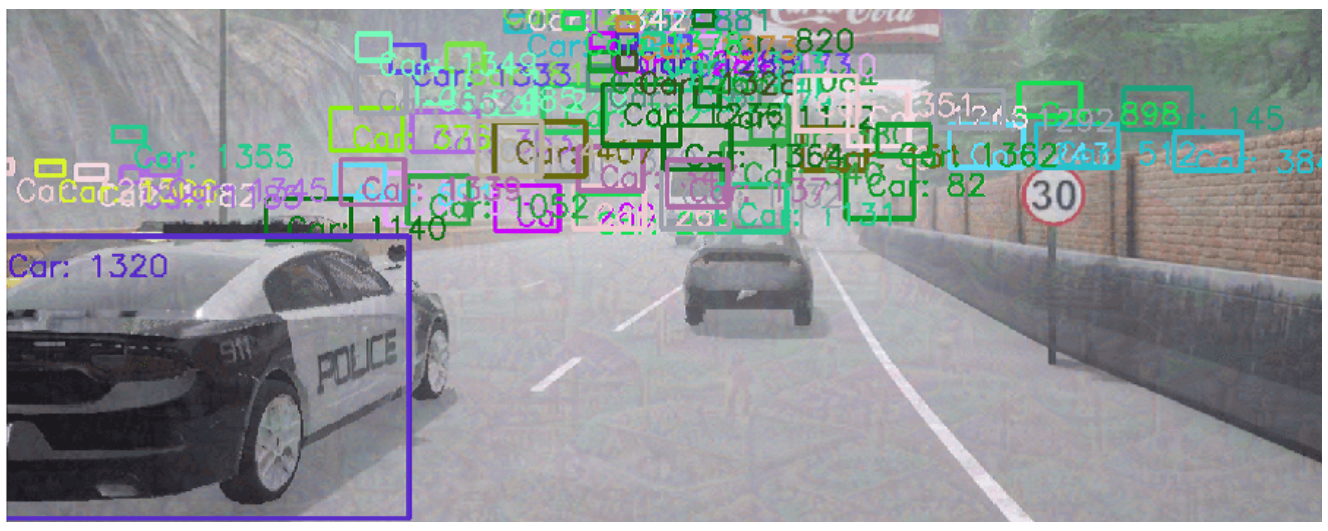
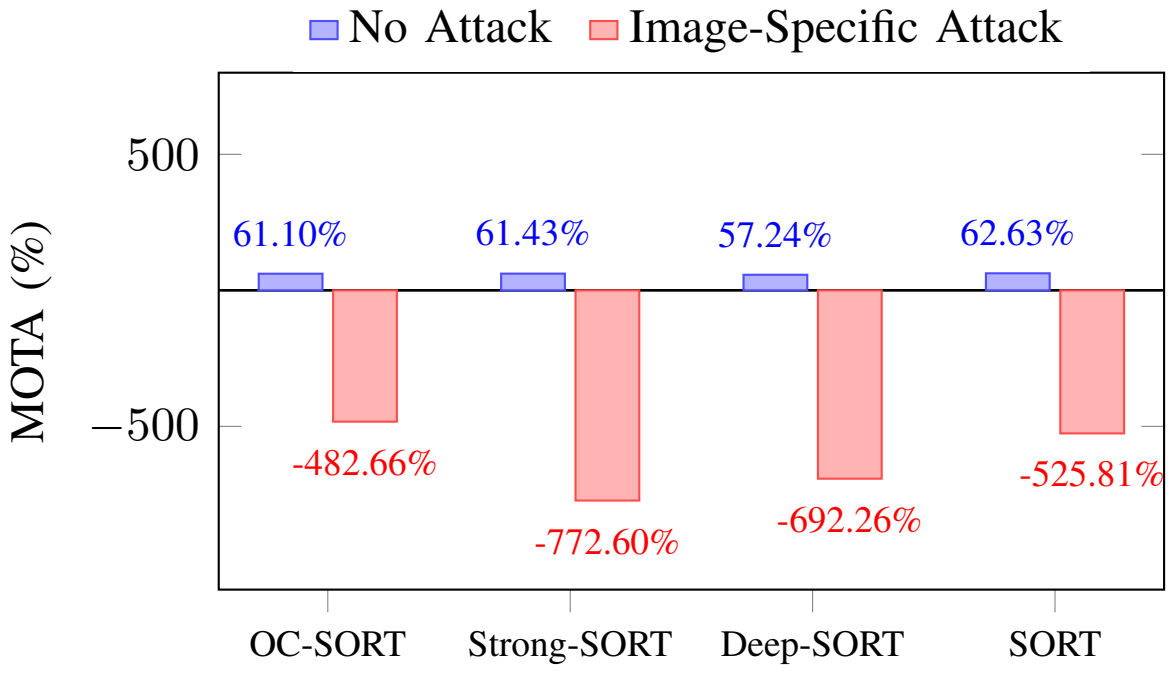
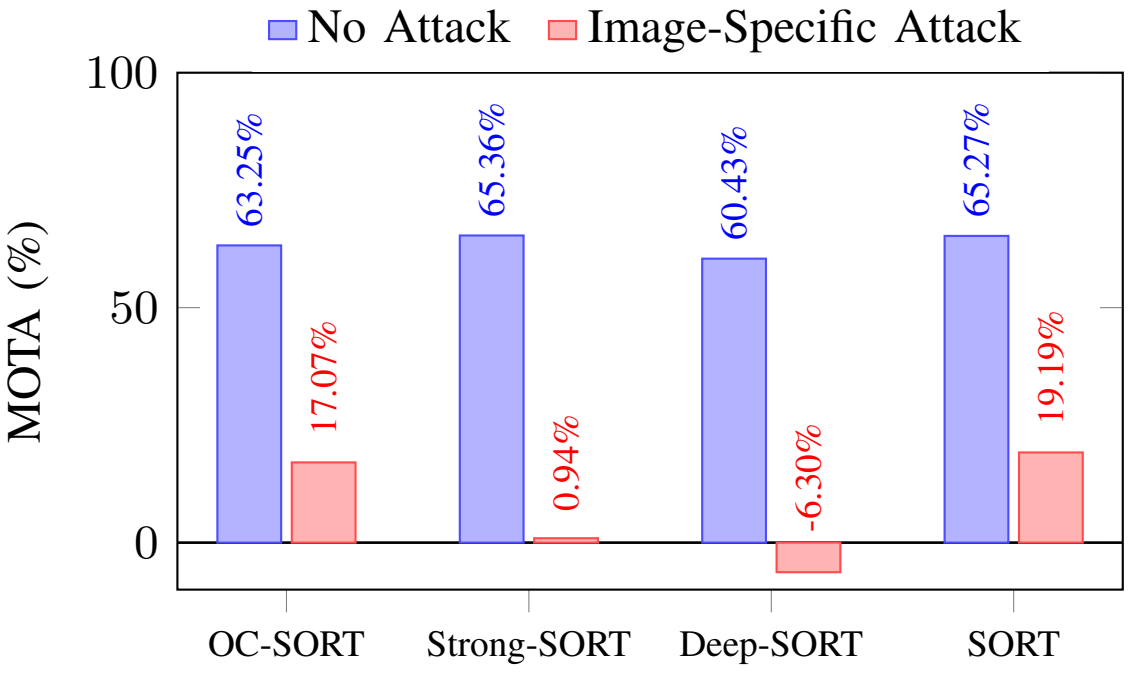
\begin{figure*}[tb]
    \centering

    \begin{subfigure}[b]{0.48\textwidth}
        \centering
        \includegraphics[
            width=\textwidth
        ]{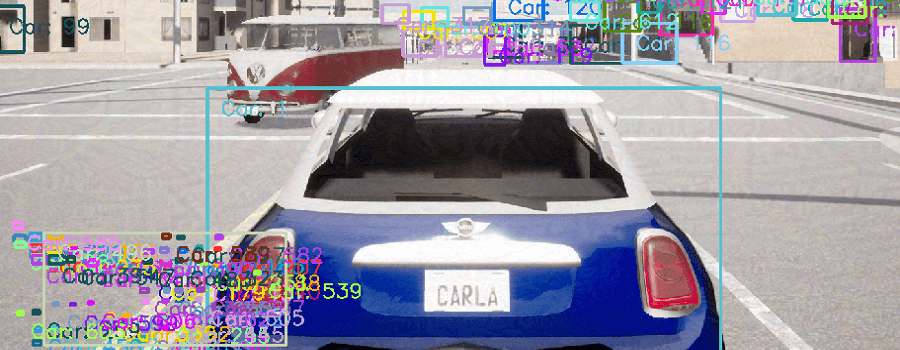}
        \caption{YOLOv3}
        \label{fig:tracking_demo_yolov3}
    \end{subfigure}
    \hfill
    \begin{subfigure}[b]{0.48\textwidth}
        \centering
        \includegraphics[
            width=\textwidth
        ]{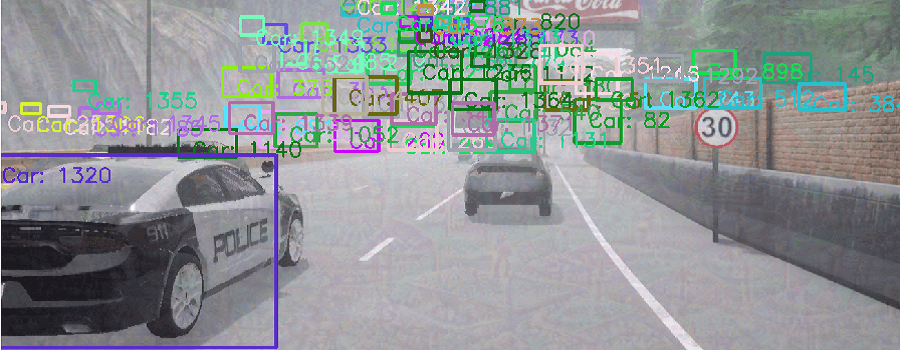}
        \caption{YOLOv4}
        \label{fig:tracking_demo_yolov4}
    \end{subfigure}

    \begin{subfigure}[b]{0.48\textwidth}
        \centering
        \begin{tikzpicture}
            \begin{axis}[
                ybar,
                enlarge x limits=0.15,
                bar width=12pt,
                width=0.9\linewidth,
                height=5.0cm,
                ymin=-1100,
                ymax=800,
                ytick={500, -500},
                ylabel={MOTA (\%)},
                symbolic x coords={
                    OC-SORT,
                    Strong-SORT,
                    Deep-SORT,
                    SORT
                },
                xtick=data,
                x tick label style={
                    rotate=0,
                    anchor=north,
                    font=\scriptsize
                },
                tick label style={font=\small},
                label style={font=\small},
                extra y ticks={0},
                extra y tick style={
                    grid=major,
                    grid style={
                        black,
                        line width=0.5pt
                    },
                    tick style={draw=none},
                    yticklabels={}
                },
                nodes near coords, 
                point meta=explicit symbolic, 
                every node near coord/.append style={ 
                    inner sep=5pt 
                }, 
                legend style={
                    font=\small,
                    at={(0.5,1.0)},
                    anchor=south,
                    legend columns=-1,
                    draw=none,
                    /tikz/every even column/.append style={
                        column sep=5pt
                    }
                },
                legend image code/.code={
                    \draw[#1,fill]
                        (0cm,-0.06cm) rectangle
                        (0.18cm,0.06cm);
                },
            ]
                \addplot[
                    nodes near coords,
                    every node near coord/.append style={
                        anchor=south,
                        font=\scriptsize,
                        color=blue,
                    },
                    fill=blue!30,
                    draw=blue!70,
                ] coordinates {
                    (OC-SORT,61.10)[61.10\%]
                    (Strong-SORT,61.43)[61.43\%]
                    (Deep-SORT,57.24)[57.24\%]
                    (SORT,62.63)[62.63\%]
                };
                \addplot [
                    nodes near coords,
                    every node near coord/.append style={
                        anchor=north,
                        font=\scriptsize,
                        color=red,
                    },
                    color=red!70,
                    fill=red!30,
                ] coordinates {
                    (OC-SORT,-482.66)[-482.66\%]
                    (Strong-SORT,-772.60)[-772.60\%]
                    (Deep-SORT,-692.26)[-692.26\%]
                    (SORT,-525.81)[-525.81\%]
                };
                \legend{No Attack, Image-Specific Attack}
            \end{axis}
        \end{tikzpicture}
        \caption{YOLOv3 MOTA}
    \end{subfigure}
    \begin{subfigure}[b]{0.48\textwidth}
        \centering
        \begin{tikzpicture}
            \begin{axis}[
                ybar,
                enlarge x limits=0.15,
                bar width=12pt,
                width=0.9\linewidth,
                height=5.0cm,
                ymin=-10,
                ymax=100,
                ylabel={MOTA (\%)},
                symbolic x coords={
                    OC-SORT,
                    Strong-SORT,
                    Deep-SORT,
                    SORT
                },
                xtick=data,
                x tick label style={
                    rotate=0,
                    anchor=north,
                    font=\scriptsize
                },
                tick label style={font=\small},
                label style={font=\small},
                extra y ticks={0},
                extra y tick style={
                    grid=major,
                    grid style={
                        black,
                        line width=0.5pt
                    },
                    tick style={draw=none},
                    yticklabels={}
                },
                nodes near coords, 
                point meta=explicit symbolic, 
                legend style={
                    font=\small,
                    at={(0.5, 1.0)},
                    anchor=south,
                    legend columns=-1,
                    draw=none,
                    /tikz/every even column/.append style={
                        column sep=5pt
                    }
                },
                legend image code/.code={
                    \draw[#1,fill]
                        (0cm,-0.06cm) rectangle
                        (0.18cm,0.06cm);
                },
            ]
                \addplot [
                    nodes near coords,
                    every node near coord/.append style={
                        anchor=west,
                        font=\scriptsize,
                        color=blue,
                        rotate=90,
                    },
                    color=blue!70,
                    fill=blue!30,
                ] coordinates {
                    (OC-SORT,63.25)[63.25\%]
                    (Strong-SORT,65.36)[65.36\%]
                    (Deep-SORT,60.43)[60.43\%]
                    (SORT,65.27)[65.27\%]
                };
                \addplot [
                    nodes near coords,
                    every node near coord/.append style={
                        anchor=west,
                        font=\scriptsize,
                        color=red,
                        rotate=90,
                        inner sep=5pt
                    },
                    color=red!70,
                    fill=red!30,
                ] coordinates {
                    (OC-SORT,17.07)[17.07\%]
                    (Strong-SORT,0.94)[0.94\%]
                    (Deep-SORT,-6.30)[ -6.30\%]
                    (SORT,19.19)[19.19\%]
                };
                \legend{No Attack, Image-Specific Attack}
            \end{axis}
        \end{tikzpicture}
        \caption{YOLOv4 MOTA}
    \end{subfigure}
    \caption{Image-Specific Attacks substantially degrade the MOTA by generating a large number of small false positives bounding boxes. In contrast, YOLOv4 produces fewer but larger boxes, making it comparatively more robust to these attacks. }
    \label{fig:tracking_specific_comparison}
\end{figure*}


\subsection{Image-Specific PCB Attack}

The image-specific attack was tested with hyperparameters of attack strength $\epsilon=8$ and learning rate decay $k=0.99$. We updated the perturbation frame-by-frame, iterating one step per frame for 300 consecutive frames on each of the 7 maps.

After applying adversarial perturbations, \acrshort{mota} decreases significantly for both YOLOv3 and YOLOv4, as shown in Fig.~\ref{fig:tracking_specific_comparison}. YOLOv3 generates many small false-positive boxes, whereas YOLOv4 produces fewer but larger spurious detections (see Fig.~\ref{fig:tracking_specific_comparison} (a) and (b)). Notably, the MOTAs of YOLOv3 drop to negative values due to a large number of false positives.




The effectiveness of the image-specific attack can be further improved by increasing the number of optimization iterations per frame, as false positives are relatively rare in the initial frames, leaving the tracking method largely unaffected. HOTA and IDF1 also decline, as shown in Appendix Figs.~\ref{fig:HOTA} and~\ref{fig:IDF1}. As a result, perturbing the object detection model alone is sufficient to compromise TBD frameworks.





\begin{figure*}[tb] 
    \centering 
 
    \begin{subfigure}[b]{0.48\textwidth} 
        \centering 
        \includegraphics[ 
            width=\textwidth 
        ]{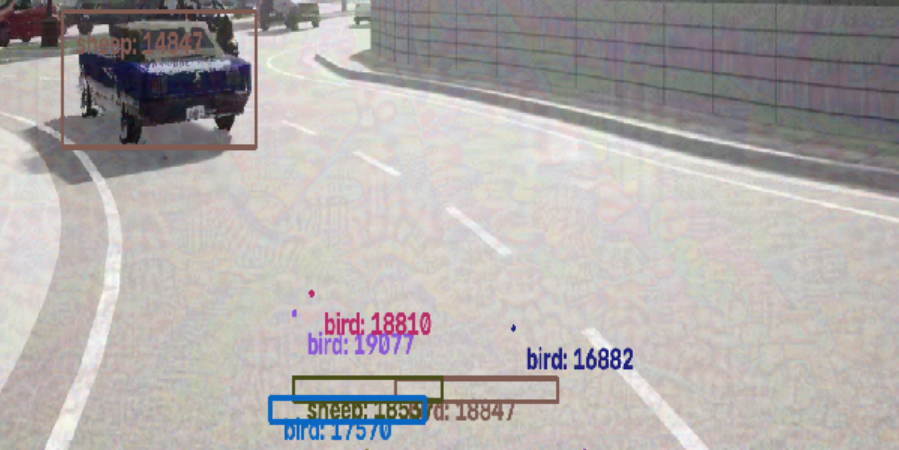} 
        \caption{YOLOv3} 
        \label{fig:tracking_yolov3} 
    \end{subfigure} 
    \begin{subfigure}[b]{0.48\textwidth} 
        \centering 
        \includegraphics[ 
            width=\textwidth 
        ]{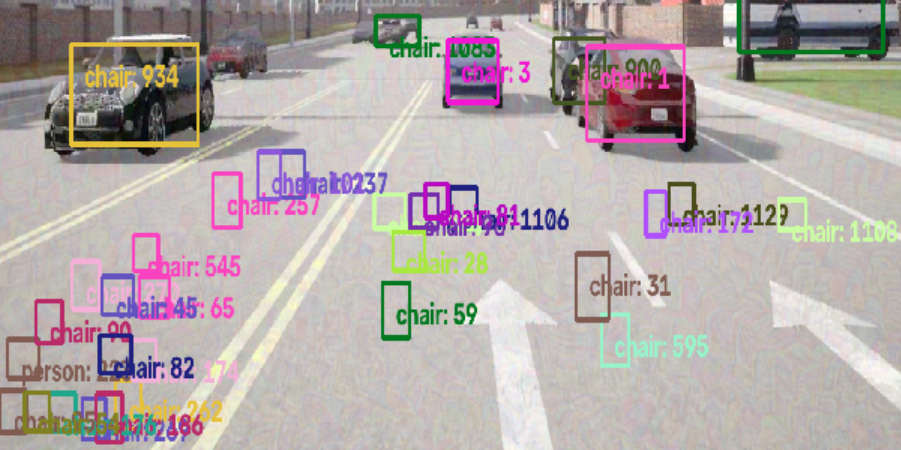} 
        \caption{YOLOv4} 
        \label{fig:tracking_yolov4} 
    \end{subfigure} 
 
    \begin{subfigure}[b]{0.48\textwidth} 
        \centering 
        \begin{tikzpicture} 
            \begin{axis}[ 
                ybar, 
                enlarge x limits=0.15,
                bar width=12pt, 
                width=0.9\linewidth, 
                height=4.5cm, 
                ymin=0, 
                ymax=100, 
                ylabel={MOTA (\%)}, 
                symbolic x coords={ 
                    OC-SORT, 
                    Strong-SORT, 
                    Deep-SORT, 
                    SORT 
                }, 
                xtick=data, 
                x tick label style={ 
                    rotate=0, 
                    anchor=north, 
                    font=\scriptsize 
                }, 
                tick label style={font=\small}, 
                label style={font=\small}, 
                nodes near coords, 
                point meta=explicit symbolic, 
                every node near coord/.append style={ 
                    anchor=west, 
                    rotate=90, 
                    font=\scriptsize, 
                    inner sep=1pt 
                }, 
                legend style={
                    font=\small,
                    at={(0.5, 1.0)},
                    anchor=south,
                    legend columns=-1,
                    draw=none,
                    /tikz/every even column/.append style={
                        column sep=5pt
                    }
                },
                legend image code/.code={
                    \draw[#1,fill]
                        (0cm,-0.06cm) rectangle
                        (0.18cm,0.06cm);
                },
            ] 
                \addplot[
                    fill=blue!30,
                    draw=blue!70,
                ] coordinates { 
                    (OC-SORT,61.10)[61.10\%] 
                    (Strong-SORT,61.43)[61.43\%] 
                    (Deep-SORT,57.24)[57.24\%] 
                    (SORT,62.63)[62.63\%] 
                }; 
 
                \addplot[
                    fill=gray!50,
                    draw=black
                ] coordinates { 
                    (OC-SORT,45.45)[45.45\%] 
                    (Strong-SORT,47.18)[47.18\%] 
                    (Deep-SORT,43.28)[43.28\%] 
                    (SORT,46.85)[46.85\%] 
                }; 
                \legend{No Attack, Image-Agnostic Attack}
            \end{axis} 
        \end{tikzpicture} 
        \caption{YOLOv3 MOTA} 
        \label{fig:tracking_uap_yolov3_mota} 
    \end{subfigure} 
    \begin{subfigure}[b]{0.48\textwidth} 
        \centering 
        \begin{tikzpicture} 
            \begin{axis}[ 
                ybar, 
                enlarge x limits=0.15,
                bar width=12pt, 
                width=0.9\linewidth, 
                height=4.5cm, 
                ymin=0, 
                ymax=100, 
                ylabel={MOTA (\%)}, 
                symbolic x coords={ 
                    OC-SORT, 
                    Strong-SORT, 
                    Deep-SORT, 
                    SORT 
                }, 
                xtick=data, 
                x tick label style={ 
                    rotate=0, 
                    anchor=north, 
                    font=\scriptsize 
                }, 
                tick label style={font=\small}, 
                label style={font=\small}, 
                nodes near coords, 
                point meta=explicit symbolic, 
                every node near coord/.append style={ 
                    anchor=west, 
                    rotate=90, 
                    font=\scriptsize, 
                    inner sep=1pt 
                }, 
                legend style={
                    font=\small,
                    at={(0.5, 1.0)},
                    anchor=south,
                    legend columns=-1,
                    draw=none,
                    /tikz/every even column/.append style={
                        column sep=5pt
                    }
                },
                legend image code/.code={
                    \draw[#1,fill]
                        (0cm,-0.06cm) rectangle
                        (0.18cm,0.06cm);
                },
            ] 
                \addplot[
                    fill=blue!30,
                    draw=blue!70
                ]
                coordinates { 
                    (OC-SORT,63.25)[63.25\%] 
                    (Strong-SORT,65.36)[65.36\%] 
                    (Deep-SORT,60.43)[60.43\%] 
                    (SORT,65.27)[65.27\%] 
                }; 
 
                \addplot[
                    fill=gray!50,
                    draw=black
                ]
                coordinates { 
                    (OC-SORT,14.26)[14.26\%] 
                    (Strong-SORT,16.95)[16.95\%] 
                    (Deep-SORT,10.98)[10.98\%] 
                    (SORT,14.70)[14.70\%] 
                };
                \legend{No Attack, Image-Agnostic Attack}
            \end{axis} 
        \end{tikzpicture} 
        \caption{YOLOv4 MOTA} 
        \label{fig:tracking_uap_yolov4_mota} 
    \end{subfigure} 
    \caption{Image-Agnostic Attacks generate a Universal Adversarial Perturbation (UAP) applied to all video frames. Although generally less effective than image-specific attacks, they can still substantially degrade MOTA.}
    \label{fig:tracking_uap_mota} 
 \end{figure*}
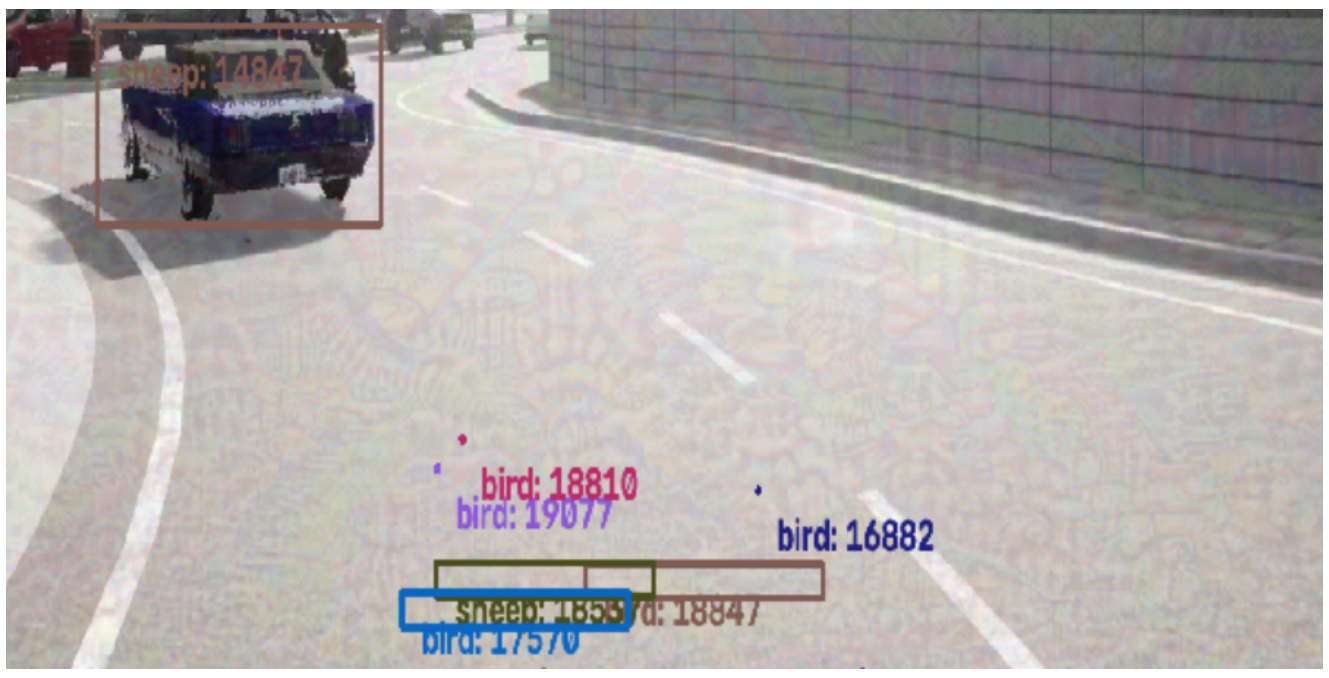
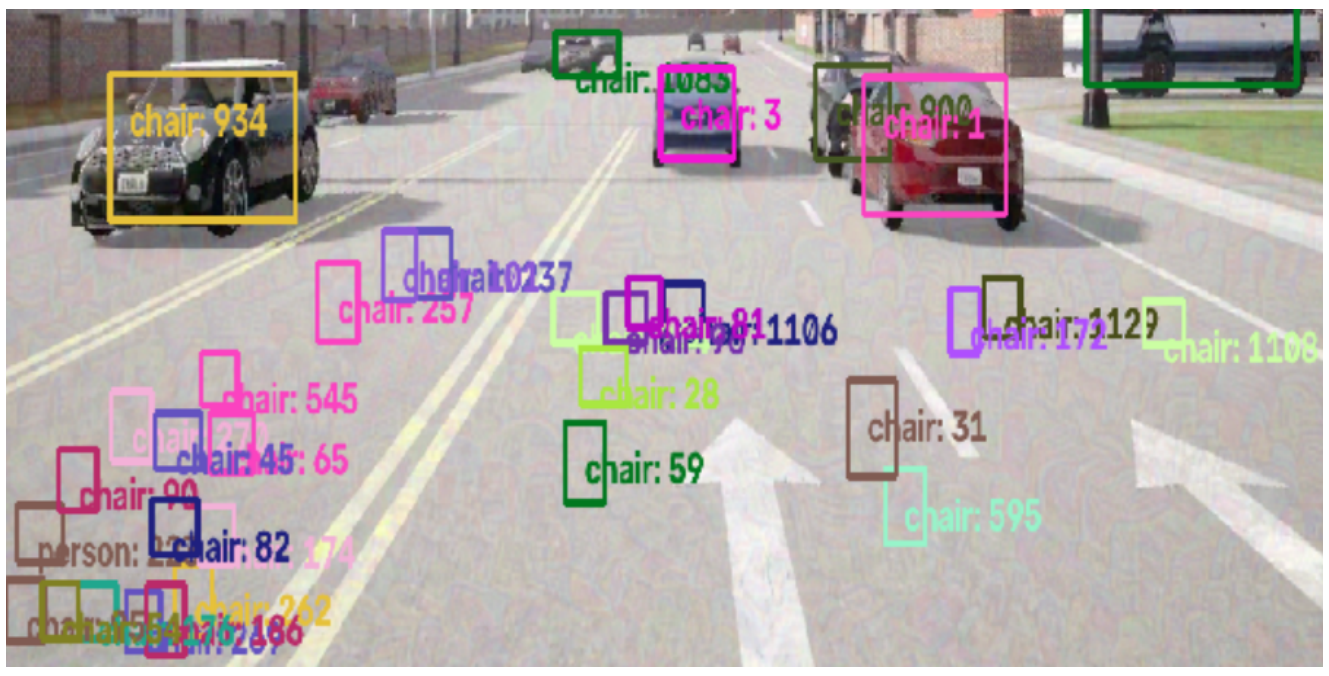
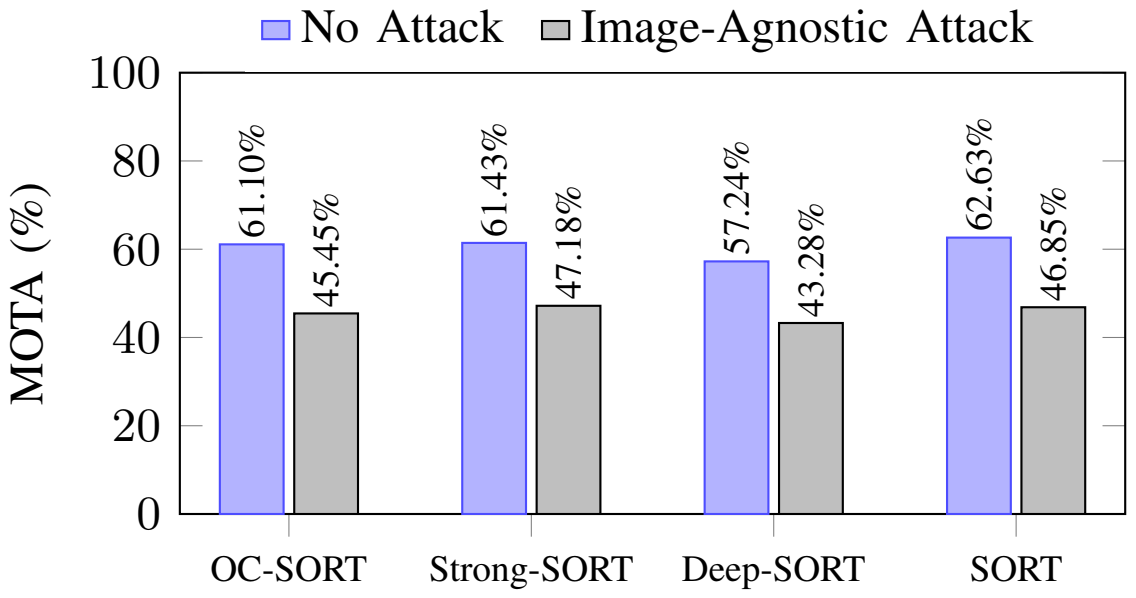
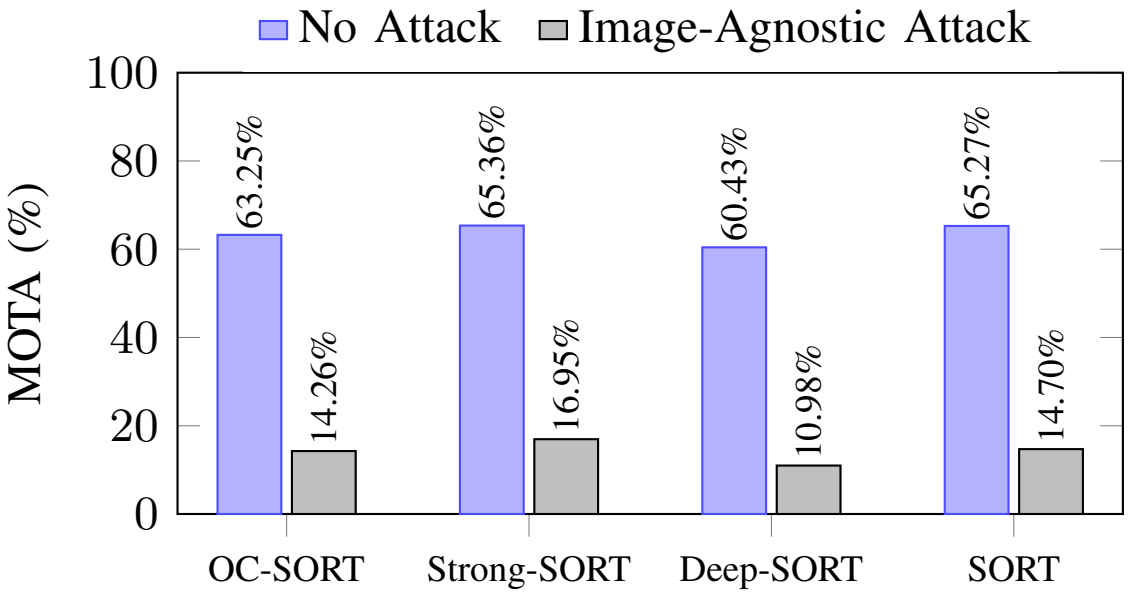

\subsection{Image-Agnostic Attack}


The image-agnostic attack generates an \acrfull{uap} to attack all video frames. We trained the \acrshort{uap} on Map 02 for 100 iterations, as Map 02 generalizes best across all maps. The training was conducted with a learning rate of $\alpha=0.001$, a learning rate decay of $k=0.98$, and an attack strength of $\epsilon=8$.

Figure~\ref{fig:tracking_uap_mota} shows that the UAP reduces MOTA for both detectors. The HOTA and IDF1 results, presented in Appendix Figs.~\ref{fig:HOTA} and~\ref{fig:IDF1}, further indicate degraded tracking performance, regarding track association and identity preservation.


Although image-agnostic attacks are generally less effective than image-specific attacks, they can still significantly reduce MOTA (see Fig. \ref{fig:tracking_uap_mota}). A key difference between attacks targeting YOLOv3 and YOLOv4 is the predicted class of false-positives: predominantly birds for YOLOv3 and chairs for YOLOv4, as shown in Fig. \ref{fig:tracking_uap_mota}(a) and (b), respectively. In autonomous driving systems, which primarily track vehicles, these small, densely distributed false-positive bounding boxes are often filtered out, reducing the effectiveness of attacks. As a result, incorporating targeted objectives into UAP generation could further improve attack effectiveness.

\section{Conclusion}

This study evaluated four real-time online tracking methods: SORT, Deep SORT, Strong SORT, and OC-SORT using the CARLA autonomous driving simulator. With the same object detector, OC-SORT achieved tracking accuracy comparable to Strong SORT without requiring additional neural networks for feature extraction. As a result, improving conventional tracking components, such as the Kalman filter and linear assignment solver, can enhance tracking accuracy without introducing additional deep learning-based components and potential attack surfaces. 


Furthermore, since the tracking performance of \acrshort{tbd} methods is highly dependent on detection accuracy, our experimental results demonstrated that both image-specific and image-agnostic attacks against object detectors alone substantially degraded \acrshort{mota}, \acrshort{hota} and \acrshort{idf1}, indicating that perturbing the object detector alone is sufficient to compromise the tracking system.

\bibliographystyle{IEEEtran}
\bibliography{references}


\appendix

\begin{figure}[ht]
    \centering
    \begin{subfigure}[b]{0.48\textwidth}
        \centering
        \begin{tikzpicture}
            \begin{axis}[
                ybar,
                enlarge x limits=0.15,
                bar width=12pt,
                width=\linewidth,
                height=4.2cm,
                ymin=0,
                ymax=100,
                ylabel={HOTA (\%)},
                symbolic x coords={
                    OC-SORT,
                    Strong-SORT,
                    Deep-SORT,
                    SORT
                },
                xtick=data,
                x tick label style={
                    rotate=0,
                    anchor=north,
                    font=\scriptsize
                },
                tick label style={font=\small},
                label style={font=\small},
                nodes near coords,
                point meta=explicit symbolic,
                every node near coord/.append style={
                    anchor=west,
                    rotate=90,
                    font=\scriptsize,
                    inner sep=1pt
                },
                legend style={
                    font=\small,
                    at={(0.5,1.03)},
                    anchor=south,
                    legend columns=-1,
                    draw=none,
                    /tikz/every even column/.append style={
                        column sep=5pt
                    }
                },
                legend image code/.code={
                    \draw[#1,fill]
                        (0cm,-0.06cm) rectangle
                        (0.18cm,0.06cm);
                },
            ]
                \addplot coordinates {
                    (OC-SORT,55.90)[55.90\%]
                    (Strong-SORT,56.19)[56.19\%]
                    (Deep-SORT,53.80)[53.80\%]
                    (SORT,53.45)[53.45\%]
                };
                \addplot coordinates {
                    (OC-SORT,17.03)[17.03\%]
                    (Strong-SORT,14.85)[14.85\%]
                    (Deep-SORT,15.32)[15.32\%]
                    (SORT,15.80)[15.80\%]
                };
                \addplot[
                    fill=gray!50,
                    draw=black
                ] 
                coordinates { 
                    (OC-SORT,43.45)[43.45\%] 
                    (Strong-SORT,43.01)[43.01\%] 
                    (Deep-SORT,43.34)[43.34\%] 
                    (SORT,40.58)[40.58\%] 
                };
                \legend{No Attack, Image-Specific, Image-Agnostic}
            \end{axis}
        \end{tikzpicture}
        \caption{YOLOv3 HOTA}
    \end{subfigure}
    \begin{subfigure}[b]{0.48\textwidth}
        \centering
        \begin{tikzpicture}
            \begin{axis}[
                ybar,
                enlarge x limits=0.15,
                bar width=12pt,
                width=\linewidth,
                height=4.2cm,
                ymin=0,
                ymax=100,
                ylabel={HOTA (\%)},
                symbolic x coords={
                    OC-SORT,
                    Strong-SORT,
                    Deep-SORT,
                    SORT
                },
                xtick=data,
                x tick label style={
                    rotate=0,
                    anchor=north,
                    font=\scriptsize
                },
                tick label style={font=\small},
                label style={font=\small},
                nodes near coords,
                point meta=explicit symbolic,
                every node near coord/.append style={
                    anchor=west,
                    rotate=90,
                    font=\scriptsize,
                    inner sep=1pt
                },
                legend style={
                    font=\scriptsize,
                    at={(0.5,1.03)},
                    anchor=south,
                    legend columns=-1,
                    draw=none,
                },
                legend image code/.code={
                    \draw[#1,fill]
                        (0cm,-0.06cm) rectangle
                        (0.18cm,0.06cm);
                },
            ]
                \addplot coordinates {
                    (OC-SORT,57.99)[57.99\%]
                    (Strong-SORT,59.49)[59.49\%]
                    (Deep-SORT,56.65)[56.65\%]
                    (SORT,56.31)[56.31\%]
                };
                \addplot coordinates {
                    (OC-SORT,25.34)[25.34\%]
                    (Strong-SORT,26.86)[26.86\%]
                    (Deep-SORT,25.06)[25.06\%]
                    (SORT,26.32)[26.32\%]
                };
                \addplot [
                    fill=gray!50,
                    draw=black
                ]
                coordinates { 
                    (OC-SORT,22.05)[22.05\%] 
                    (Strong-SORT,24.11)[24.11\%] 
                    (Deep-SORT,24.87)[24.87\%] 
                    (SORT,21.96)[21.96\%] 
                }; 
            \end{axis}
        \end{tikzpicture}
        \caption{YOLOv4 HOTA}
    \end{subfigure}
    \caption{HOTA under Image-Specific and Image-Agnostic Attacks against YOLOv3 and YOLOv4.}
    \label{fig:HOTA}
\end{figure}
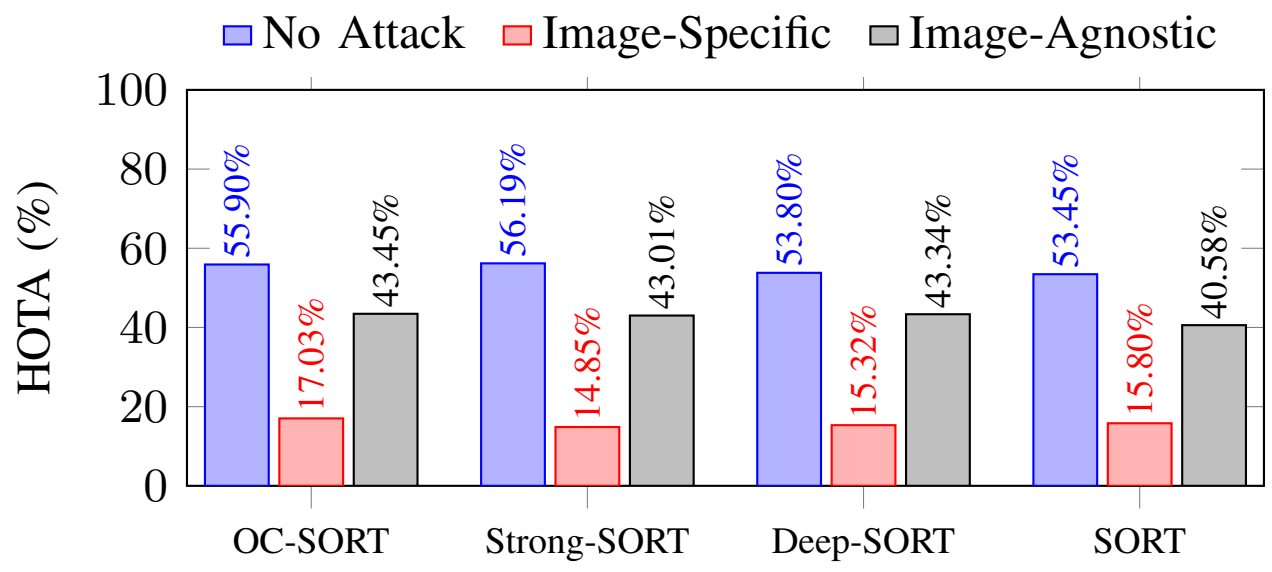
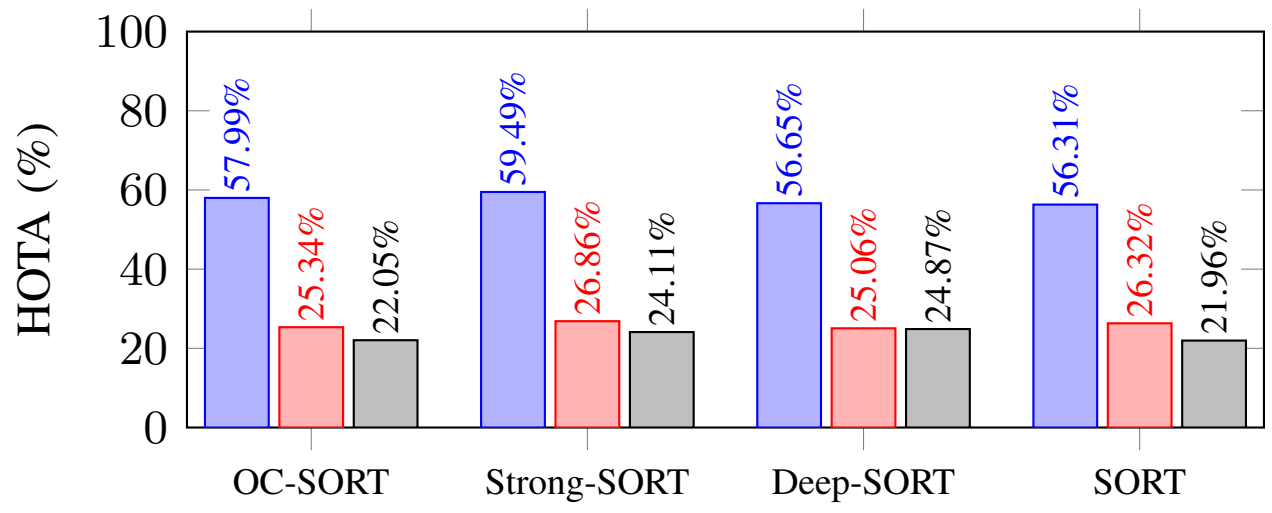

\begin{figure}[ht] 
    \centering 
    \begin{subfigure}[b]{0.48\textwidth}
        \centering
        \begin{tikzpicture}
            \begin{axis}[
                ybar,
                enlarge x limits=0.15,
                bar width=12pt,
                width=\linewidth,
                height=4.6cm,
                ymin=0,
                ymax=110,
                ylabel={IDF1 (\%)},
                symbolic x coords={
                    OC-SORT,
                    Strong-SORT,
                    Deep-SORT,
                    SORT
                },
                xtick=data,
                x tick label style={
                    rotate=0,
                    anchor=north,
                    font=\scriptsize
                },
                tick label style={font=\small},
                label style={font=\small},
                nodes near coords,
                point meta=explicit symbolic,
                every node near coord/.append style={
                    anchor=west,
                    rotate=90,
                    font=\scriptsize,
                    inner sep=1pt
                },
                legend style={
                    font=\small,
                    at={(0.5,1.03)},
                    anchor=south,
                    legend columns=-1,
                    draw=none,
                    /tikz/every even column/.append style={
                        column sep=5pt
                    }
                },
                legend image code/.code={
                    \draw[#1,fill]
                        (0cm,-0.06cm) rectangle
                        (0.18cm,0.06cm);
                },
            ]
                \addplot coordinates {
                    (OC-SORT,74.31)[74.31\%]
                    (Strong-SORT,71.43)[71.43\%]
                    (Deep-SORT,70.92)[70.92\%]
                    (SORT,69.30)[69.30\%]
                };
                \addplot coordinates {
                    (OC-SORT,14.09)[14.09\%]
                    (Strong-SORT,9.95)[9.95\%]
                    (Deep-SORT,11.29)[11.29\%]
                    (SORT,12.00)[12.00\%]
                };
                \addplot[
                    fill=gray!50,
                    draw=black
                ]
                coordinates { 
                    (OC-SORT,62.43)[62.43\%] 
                    (Strong-SORT,57.42)[57.42\%] 
                    (Deep-SORT,60.17)[60.17\%] 
                    (SORT,55.20)[55.20\%] 
                };
                \legend{No Attack, Image-Specific, Image-Agnostic}
            \end{axis}
        \end{tikzpicture}
        \caption{YOLOv3 IDF1}
    \end{subfigure}
    \begin{subfigure}[b]{0.48\textwidth}
        \centering
        \begin{tikzpicture}
            \begin{axis}[
                ybar,
                enlarge x limits=0.15,
                bar width=12pt,
                width=\linewidth,
                height=4.6cm,
                ymin=0,
                ymax=110,
                ylabel={IDF1 (\%)},
                symbolic x coords={
                    OC-SORT,
                    Strong-SORT,
                    Deep-SORT,
                    SORT
                },
                xtick=data,
                x tick label style={
                    rotate=0,
                    anchor=north,
                    font=\scriptsize
                },
                tick label style={font=\small},
                label style={font=\small},
                nodes near coords,
                point meta=explicit symbolic,
                every node near coord/.append style={
                    anchor=west,
                    rotate=90,
                    font=\scriptsize,
                    inner sep=1pt
                },
                legend style={
                    font=\small,
                    at={(0.5,1.03)},
                    anchor=south,
                    legend columns=-1,
                    draw=none,
                    /tikz/every even column/.append style={
                        column sep=5pt
                    }
                },
                legend image code/.code={
                    \draw[#1,fill]
                        (0cm,-0.06cm) rectangle
                        (0.18cm,0.06cm);
                },
            ]
                \addplot coordinates {
                    (OC-SORT,75.78)[75.78\%]
                    (Strong-SORT,74.81)[74.81\%]
                    (Deep-SORT,74.07)[74.07\%]
                    (SORT,73.37)[73.37\%]
                };
                \addplot coordinates {
                    (OC-SORT,34.46)[34.46\%]
                    (Strong-SORT,33.20)[33.20\%]
                    (Deep-SORT,32.95)[32.95\%]
                    (SORT,34.78)[34.78\%]
                };
                \addplot[
                    fill=gray!50,
                    draw=black
                ]
                coordinates { 
                    (OC-SORT,33.07)[33.07\%] 
                    (Strong-SORT,34.37)[34.37\%] 
                    (Deep-SORT,33.54)[33.54\%] 
                    (SORT,30.20)[30.20\%] 
                }; 
            \end{axis}
        \end{tikzpicture}
        \caption{YOLOv4 IDF1}
    \end{subfigure}
    \caption{IDF1 under Image-Specific and Image-Agnostic Attacks against YOLOv3  and YOLOv4.}
    \label{fig:IDF1}
\end{figure}
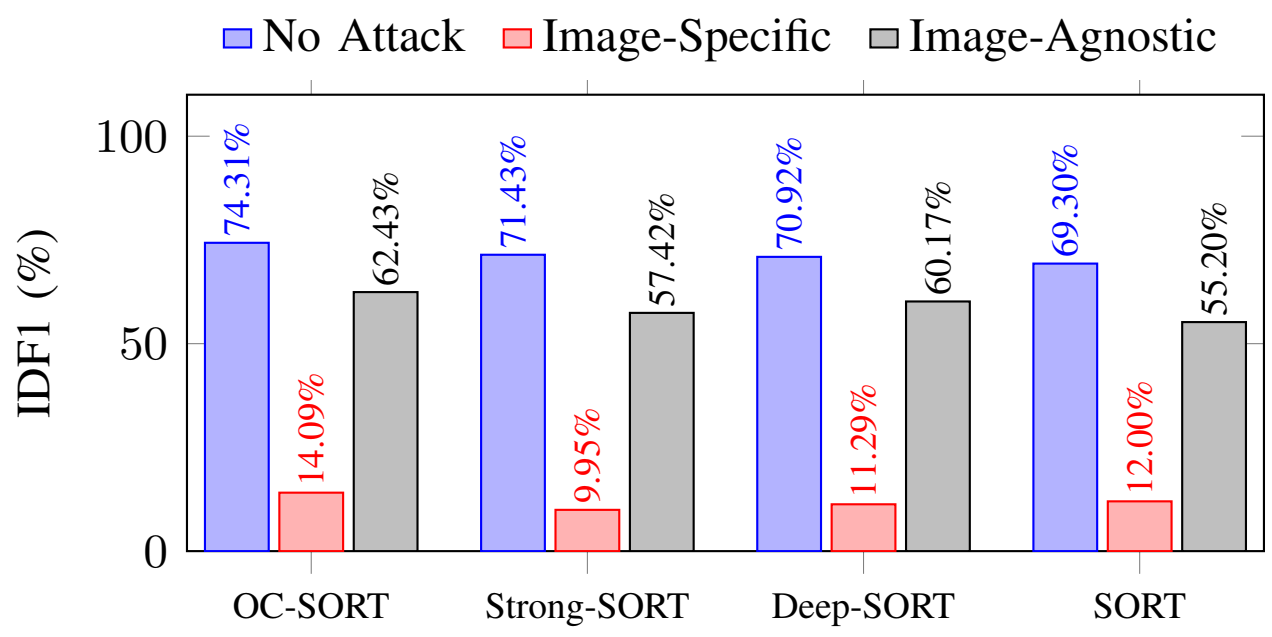
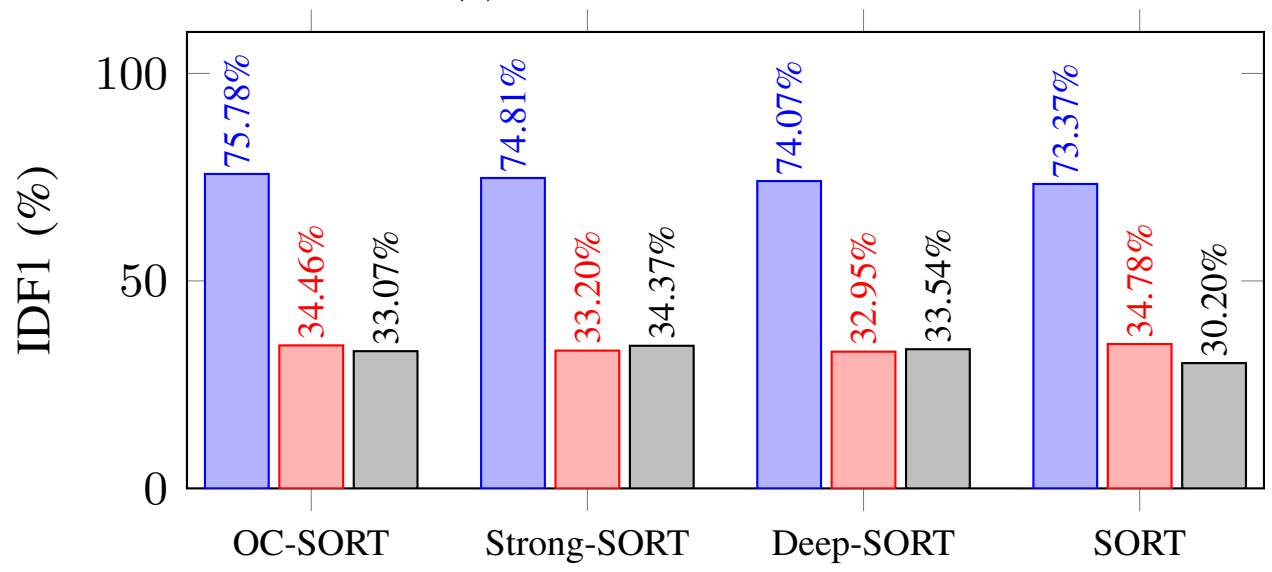

\end{document}